\documentclass[11pt]{article}
\usepackage[a4paper,margin=0.78in]{geometry}
\usepackage[T1]{fontenc}
\usepackage[utf8]{inputenc}
\usepackage{graphicx}
\usepackage{booktabs}
\usepackage{tabularx}
\usepackage{array}
\usepackage{amsmath}
\usepackage{amssymb}
\usepackage{siunitx}
\usepackage{caption}
\usepackage[table]{xcolor}
\usepackage{colortbl}
\definecolor{E3HeadSoft}{HTML}{C9DCFB}
\definecolor{E3CellSoft}{HTML}{EAF1FC}
\definecolor{DeltaUp}{HTML}{1E7A3C}
\definecolor{DeltaDown}{HTML}{B23A3A}
\definecolor{MetricStripe}{HTML}{F4F7FC}
\newcommand{\deltaup}[1]{\textcolor{DeltaUp}{$\uparrow$\,#1\,pp}}

\usepackage{float}
\usepackage{microtype}
\usepackage[round,authoryear]{natbib}
\usepackage[hidelinks=false,colorlinks=true,linkcolor=red,citecolor=red,urlcolor=red,breaklinks=true]{hyperref}
\hypersetup{linktoc=all}
\DeclareUnicodeCharacter{2501}{-}
\DeclareUnicodeCharacter{2013}{--}
\newcommand{\system}{E3}

\newcommand{\caught}{\textsc{Caught}}
\newcommand{\partialhit}{\textsc{Partial}}
\newcommand{\missed}{\textsc{Missed}}
\date{}

\begin{document}

\begin{center}
\vspace*{-0.15em}
\begin{tabular}{@{}>{\centering\arraybackslash}m{1.55cm} @{\hspace{0.45em}} >{\centering\arraybackslash}p{0.76\linewidth}@{}}
\includegraphics[width=1.35cm]{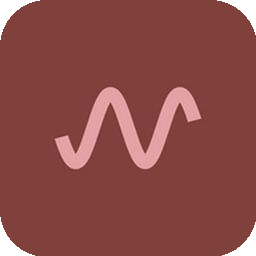} &
{\LARGE\bfseries \system{}: Issue-Level Backtesting for Automated Research Critique} \\[0.35em]
& {\normalsize A Technical Analysis on 100 ICLR 2026 Papers} \\[0.85em]
& {\large Yashwardhan Chaudhuri \quad Sanyam Jain \quad Paridhi Mundra} \\[0.25em]
& {\texttt{\{yashwardhan, sanyam, paridhi\}@noteweave.io}}
\end{tabular}
\end{center}
\vspace{0.35em}

\begin{abstract}
We present \system{}, an automated review assistant that augments reviewers and engineering teams by identifying decision-relevant technical concerns in research papers. For each concern, \system{} reports its nature, its location, its bearing on the contribution, and the analysis or evidence that would resolve it, covering unsupported claims, missing ablations, weak baselines, hidden assumptions, threats to validity, and leakage risks. To evaluate \system{} without contamination confounds we adopt an issue-level backtesting protocol: the corpus is restricted to papers postdating the training cutoff of every automated source, and for each paper a meta-judge that observes only anonymised reviews labels every (issue, source) pair as \caught{}, \partialhit{}, or \missed{}. Applied to 100 ICLR 2026 papers (4,598 judged issue rows), comparing \system{} against the ICLR human reviews and two prompt-matched LLM baselines built on \texttt{gpt-5.4} (OpenAI) and \texttt{claude-opus-4-6} (Anthropic), with meta-judge \texttt{gpt-5.5}: \textbf{\system{} attains the highest recall on every aggregate metric}. Partial-inclusive recall reaches \textbf{90.2\%} (\deltaup{15.5} over GPT, \deltaup{17.1} over Claude, \deltaup{29.2} over the human reviews) and strict recall preserves the ordering at \textbf{65.8\%}. On concerns raised by the human reviewers \system{} recovers \textbf{89.6\%}; on concerns the human reviewers missed it surfaces \textbf{1,635} additional rows admitted into the judged union, 406 above the next-best source. Corpus, baseline prompts, judge prompt template, and evaluation code are released.
\end{abstract}

\section{Introduction}
Peer review is the principal quality-control mechanism for machine-learning publishing, and its capacity is bounded by the supply of qualified reviewers. The NeurIPS consistency experiments establish that even careful reviewing is statistically noisy: when two independent committees reviewed the same submissions in 2014 and again in 2021, they disagreed on roughly a quarter of accept/reject decisions and would have reversed about half of the accept list on a rerun \citep{cortes2021inconsistency,beygelzimer2023consistency}. The major venues now receive on the order of $10^{4}$ submissions per cycle while the reviewer pool grows at a much slower rate, compressing the per-paper time budget. As a consequence, concerns that ought to be surfaced by a careful read (a missing ablation, an unsupported claim, a leakage risk, an unfair baseline) are increasingly missed not for subtlety but for lack of reviewer bandwidth, and corpus-level analyses suggest that 6.5--16.9\% of text in recent ICLR, NeurIPS, CoRL, and EMNLP reviews is already being substantially modified by LLMs in practice \citep{liang2024monitoring}.

We present \system{}, an automated review assistant that addresses this bottleneck. For a given paper, \system{} produces a structured technical critique in which each decision-relevant concern is articulated as a discrete entry: its nature, its location, its bearing on the contribution, and the analysis or evidence that would resolve it. The concerns targeted include unsupported claims, missing ablations, weak baselines, hidden assumptions, threats to validity, and leakage risks \citep{kapoor2023leakage}. \system{} is positioned as augmentation rather than substitution: it accelerates the location and articulation of hidden technical faults, so that reviewer attention can concentrate on the judgements only humans can make. The same workflow applies to engineering teams running internal audits on in-progress research.

Evaluating assistants of this kind requires care: as preprints and public OpenReview content are absorbed into LLM pretraining corpora, apparently high recall on a familiar paper may reflect prior exposure rather than analysis \citep{white2024livebench}. We therefore grade \system{} with an \emph{issue-level backtesting} protocol: (i) the corpus is restricted to papers postdating the training cutoff of every model used in the protocol; (ii) for each paper we construct a union of distinct concerns surfaced by any reviewer (human or automated) and score each (issue, source) pair separately as \caught{}, \partialhit{}, or \missed{}; (iii) labels are produced by a meta-judge that observes the four review streams in anonymised form \citep{zheng2023mtbench}.

Applied to 100 ICLR 2026 papers (4,598-row judged union, mean 46.0 concerns per paper), the protocol compares \system{} against the ICLR human reviews and two LLM baselines that run a released flaw-finding prompt on \texttt{gpt-5.4} (OpenAI) and \texttt{claude-opus-4-6} (Anthropic); the meta-judge is \texttt{gpt-5.5}. \textbf{\system{} attains the highest recall on every aggregate metric}, with partial-inclusive recall \textbf{90.2\%} (\deltaup{15.5} over GPT, \deltaup{17.1} over Claude, \deltaup{29.2} over the human reviews) and strict recall preserving the ordering. The remainder of the paper formalises the protocol (Section~\ref{sec:design}), defines the metrics (Section~\ref{sec:metrics}), reports per-source coverage (Sections~\ref{sec:results}--\ref{sec:alignment}), characterises residual errors by taxonomy (Sections~\ref{sec:taxonomy}--\ref{sec:complementarity}), and states limitations (Section~\ref{sec:limitations}).

\paragraph{Contributions.}
\begin{enumerate}\itemsep0.1em
\item \system{}, an automated review assistant that produces a structured technical critique of a research paper in which each decision-relevant concern is articulated as a discrete, reasoned entry (nature, location, bearing on the contribution, and path to resolution); intended to augment human reviewers and engineering teams by accelerating the search for hidden technical faults at scale.
\item An issue-level backtesting protocol for unbiased evaluation of critique systems, combining a post-cutoff corpus, a per-paper concern union, and a blinded meta-judge that separates critique generation from critique evaluation.
\item A like-for-like comparison of \system{} against the ICLR human reviews and prompt-matched GPT and Claude baselines on 4,598 judged issue rows from 100 ICLR 2026 papers, with per-source measurements of strict and partial-inclusive recall, weighted coverage, best-rigour share, recall on the human-salient subset, and the volume of human-missed concerns recovered, stratified by issue severity, decision tier, and taxonomy.
\item Public release of the baseline flaw-finder prompt and the meta-judge prompt template (Appendices~\ref{app:vanilla-prompt}--\ref{app:judge-prompt}).
\end{enumerate}

\section{Background and Motivation}\label{sec:related}
Three established facts motivate the system and the protocol. Unaided peer review at ML venues is statistically noisy: the two NeurIPS consistency experiments measured a 23--26\% disagreement rate between independent committees on accept/reject recommendations, with roughly half of the accept list flipping on a rerun \citep{cortes2021inconsistency,beygelzimer2023consistency}. The issue categories \system{} targets (leakage, weak baselines, hidden assumptions) are the same ones independent audits of ML-based science identify as systematically under-caught at publication time \citep{kapoor2023leakage}. LLM participation in reviewing is already underway, with corpus-level analyses estimating that 6.5--16.9\% of text in recent ICLR, NeurIPS, CoRL, and EMNLP reviews has been substantially modified by an LLM \citep{liang2024monitoring}. The protocol below answers how to measure such a system's contribution rigorously; it inherits blinded LLM-as-judge scoring from MT-Bench / Chatbot Arena \citep{zheng2023mtbench} and a post-training-cutoff corpus from contamination-aware benchmarking \citep{white2024livebench}, both adapted to the review-critique setting.

\section{Evaluation Design}\label{sec:design}
\subsection{Corpus}
We use 100 ICLR 2026 papers with publicly available reviews, stratified by decision folder: 28 oral, 28 accepted/poster, 15 conditional, and 29 rejected. The judged union holds 4,598 issue rows, averaging 46.0 per paper (median 45.5).

\begin{figure}[!ht]
\centering
\includegraphics[width=\linewidth]{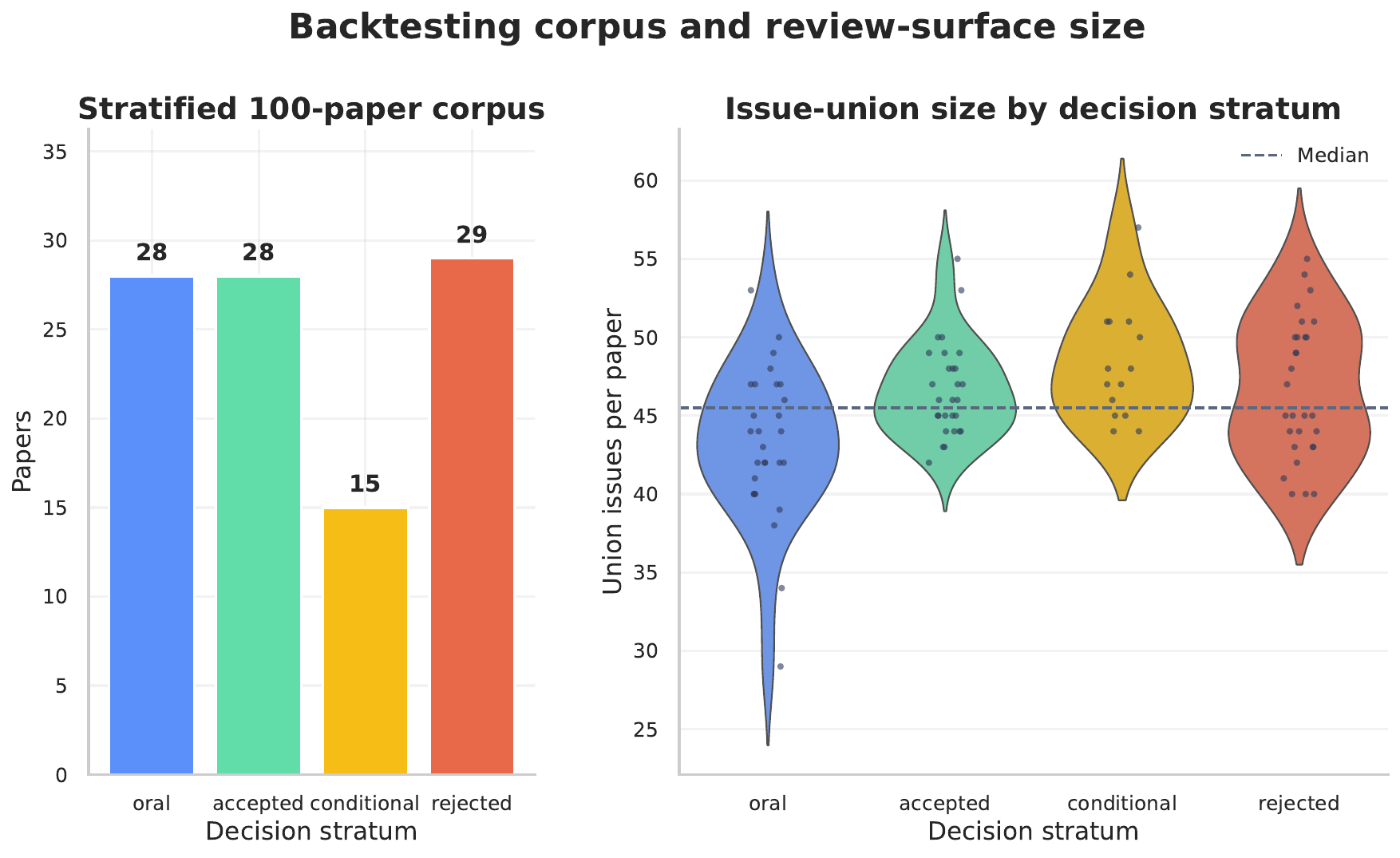}
\caption{\textbf{Corpus composition and review-surface size.} \textit{Left}: paper count per ICLR 2026 decision stratum (28 oral, 28 accepted/poster, 15 conditional, 29 rejected). \textit{Right}: violin of judged issue rows per paper; each black dot is one paper and the dashed line is the corpus median (45.5 issues). Rejected papers carry the widest issue surface and therefore the toughest recall test, which motivates the decision-stratified analysis in Section~\ref{sec:results}.}
\label{fig:dataset}
\end{figure}

Figure~\ref{fig:dataset} fixes two corpus facts before any analysis. No single decision tier dominates the sample (left). Issue-row counts vary widely per paper (a paper with 55 rows has more places to miss than one with 35), so this distribution sets the difficulty floor for recall.

\subsection{Sources Compared}
Four review streams feed the protocol per paper. \emph{Human} bundles the publicly available ICLR review material: individual reviews, the meta-review, the decision record, and associated metadata. \emph{\system{}} is the automated critique output. \emph{GPT} and \emph{Claude} are flaw-finder baselines running \texttt{gpt-5.4} (OpenAI) and \texttt{claude-opus-4-6} (Anthropic) on the paper text. Both use the same flaw-finder prompt (Appendix~\ref{app:vanilla-prompt}), which targets unsupported claims, hidden assumptions, weak baselines, confounds, missing validations, and domain-specific risks.

\subsection{Blinded Issue Judge}
A \emph{meta-judge} is a separate model call that scores the four reviews. We use \texttt{gpt-5.5} (OpenAI), distinct from every review source. The judge sees four anonymised streams (M1--M4) and never learns which is which. For every paper it (i) extracts each stream's concerns, (ii) merges them into one union (only genuinely identical concerns merged), and (iii) labels each row with a severity, a per-source status (\caught{}/\partialhit{}/\missed{}), and a \emph{best-rigour} tag for the source that addressed the issue most thoroughly. The exact prompt is in Appendix~\ref{app:judge-prompt}.

Separating critique from judgment stops any source from grading itself. Blinding stops the judge from leaning on provider names. We remap M1--M4 to source names only after the JSON validates.

\subsection{Backtesting Protocol}
\emph{Backtesting} here means evaluating critique sources on papers that postdate the LLM training cutoffs. ICLR 2026 submissions opened in late 2025 and the public reviews were released after that, which is later than the cutoffs of \texttt{gpt-5.4}, \texttt{claude-opus-4-6}, and \texttt{gpt-5.5}. The LLM baselines and judge therefore cannot earn a \caught{} from memorisation; every catch comes from reading the paper.

The protocol itself is one pass per paper: collect the four reviews (Human, \system{}, GPT, Claude), pass them anonymised through the blinded judge, and obtain one issue union with per-source statuses. All sources are scored on the same rows, so recall numbers are directly comparable. The grading unit (an issue row) is also the unit of review utility: a review helps when it names a concrete concern. Every input is public, so the evaluation can be rerun end-to-end.

\section{Metrics}\label{sec:metrics}
For every (issue, source) pair the judge writes one of three labels: \caught{}, \partialhit{}, or \missed{}. It also tags one source per issue as having the most thorough treatment: the \emph{best-rigour} tag. Six numbers summarise each source $M$ over the union of $N$ judged issues. Table~\ref{tab:metric-definitions} gives the definitions; we use the same symbols throughout the paper.

\paragraph{Notation.} We write $\#\caught{}$ for the number of issues labelled \caught{} for source $M$, and likewise $\#\partialhit{}$ and $\#\missed{}$. More generally, $\#\{i:\,\text{condition}\}$ denotes the number of issues $i$ satisfying that condition. $\mathcal{H}$ denotes the \emph{human-salient} set: issues the Human source caught or partially caught. $|\mathcal{H}|$ is its size. ``$M$ hit $i$'' is shorthand for status \caught{} or \partialhit{}.

\begin{table}[!ht]
\centering
\small
\renewcommand{\arraystretch}{1.35}
\begin{tabularx}{\linewidth}{@{}p{4.3cm} l X@{}}
\toprule
\textbf{Metric} & \textbf{Formula} & \textbf{Plain reading} \\
\midrule
\rowcolor{MetricStripe}
Strict recall, $R_{\mathrm{strict}}(M)$ &
$\dfrac{\#\caught}{N}$ &
Share of issues $M$ caught outright. \\
Partial-inclusive recall, $R_{\mathrm{hit}}(M)$ &
$\dfrac{\#\caught + \#\partialhit}{N}$ &
Share of issues $M$ raised at all (full or partial). \\
\rowcolor{MetricStripe}
Weighted coverage, $C_w(M)$ &
$\dfrac{\#\caught + 0.5\cdot\#\partialhit}{N}$ &
Partial mentions count half. \\
Best-rigour share, $B(M)$ &
$\dfrac{\#\{i:\, b_i = M\}}{N}$ &
Share of rows where the judge ranked $M$'s treatment most thorough. \\
\rowcolor{MetricStripe}
Agreement with humans, $R_{M\mid H}$ &
$\dfrac{\#\{i \in \mathcal{H}:\, M \text{ hit } i\}}{|\mathcal{H}|}$ &
Recall on the slice humans themselves raised. \\
Value beyond humans, $D_{M,\bar H}$ &
$\#\{i \notin \mathcal{H}:\, M \text{ hit } i\}$ &
Count of human-missed issues $M$ still catches. \\
\bottomrule
\end{tabularx}
\caption{Metric definitions. Counts are over the union of $N$ judged issue rows for source $M$. $B(M)$ closes a loophole in $R_{\mathrm{hit}}$: a source cannot inflate it by mentioning issues only in passing, because the best-rigour tag requires the most detailed, evidence-backed treatment on a row. The six metrics answer four questions: how often $M$ raises a real concern ($R_{\mathrm{hit}}$, $R_{\mathrm{strict}}$, $C_w$), how often its treatment is the most thorough ($B$), how well it tracks humans ($R_{M\mid H}$), and how much it adds beyond them ($D_{M,\bar H}$).}
\label{tab:metric-definitions}
\end{table}

\section{Results}\label{sec:results}
\subsection{Per-Source Coverage Across the Full Issue Union}
\begin{figure}[!ht]
\centering
\includegraphics[width=\linewidth]{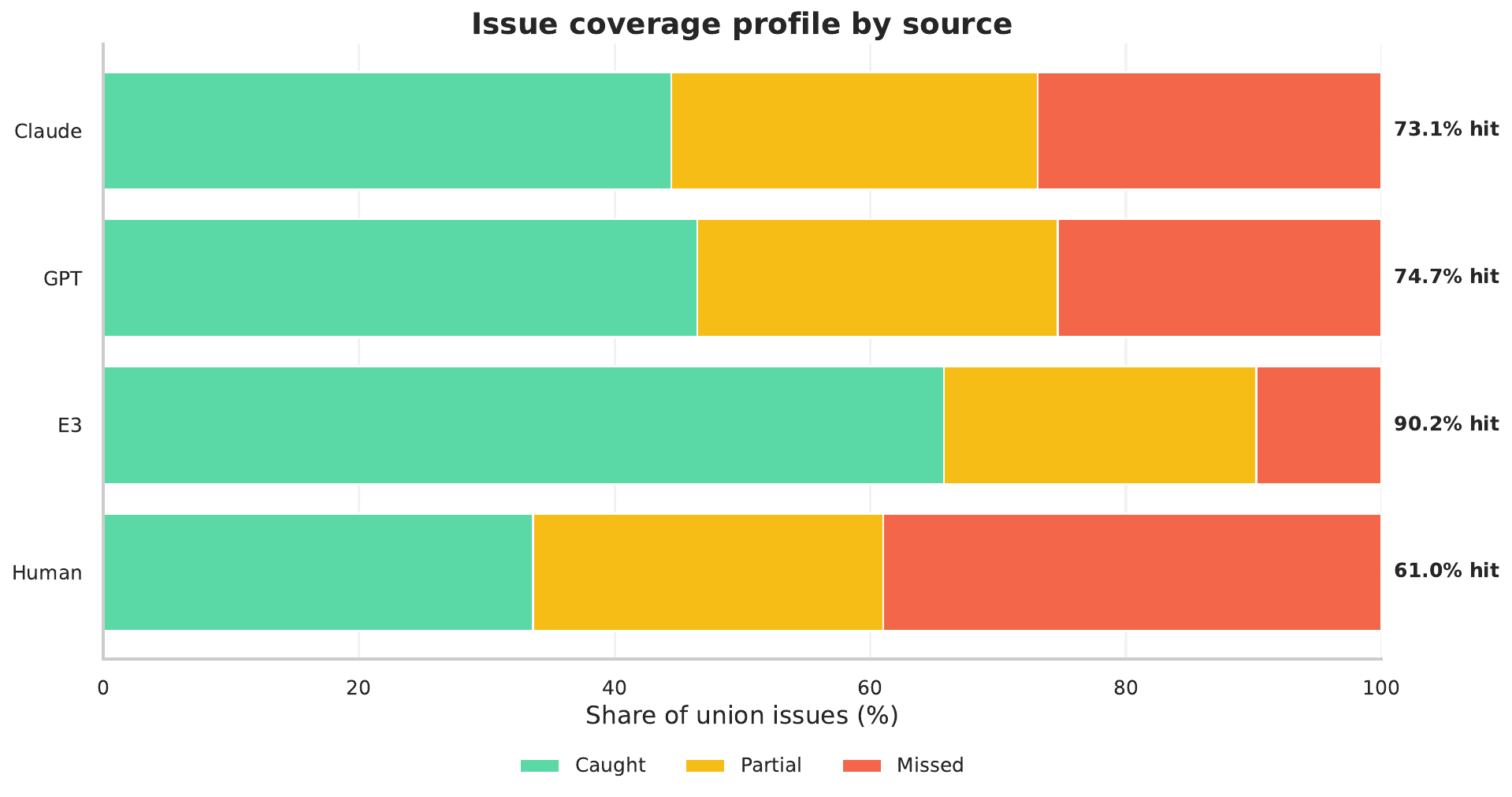}
\caption{\textbf{Coverage per source over the 4,598-row judged union.} Each bar splits one source into \caught{} (green), \partialhit{} (amber), and \missed{} (red); the right-edge label is partial-inclusive recall $R_{\mathrm{hit}}$. Two facts read off the figure directly: green+amber length orders the sources \system{} $>$ GPT $>$ Claude $>$ Human, and only the Human source has \missed{} as its largest segment, so the union contains many machine-surfaced concerns the human reviewers did not raise. \textit{Numerical companion}: Table~\ref{tab:method-summary}.}
\label{fig:coverage}
\end{figure}

Figure~\ref{fig:coverage} splits every source's 4,598-row record into \caught{}, \partialhit{}, and \missed{}; Table~\ref{tab:method-summary} prints the matching numbers along with weighted coverage $C_w$ (partial counts half) and best-rigour share $B$ (the source the judge ranked most thorough on each row). \textbf{\system{} leads every column} of Table~\ref{tab:method-summary}: partial-inclusive recall \textbf{90.2\%} (\deltaup{15.5} over GPT, \deltaup{17.1} over Claude, \deltaup{29.2} over humans), strict recall \textbf{65.8\%} (\deltaup{19.3} over GPT, \deltaup{21.3} over Claude, \deltaup{32.1} over humans), weighted coverage \textbf{78.0}, and best-rigour share \textbf{48.5\%} (\deltaup{29.9} over the best non-\system{} source). The strict-recall margin is the most demanding comparison: it only credits outright catches, and \system{} keeps the lead by at least 19.3\,pp.

\begin{table}[!ht]
\centering
\small
\begin{tabular}{lrrrrrr}
\toprule
\textbf{Source} & \textbf{Caught} & \textbf{Partial} & \textbf{Missed} & \textbf{Hit \%} & \textbf{Weighted score} & \textbf{Best-rigour \%} \\
\midrule
Human & 1,546 & 1,259 & 1,793 & 61.0 & 47.3 & 18.5 \\
\cellcolor{E3CellSoft}E3 & \cellcolor{E3CellSoft}\textbf{3,024} & \cellcolor{E3CellSoft}1,123 & \cellcolor{E3CellSoft}\textbf{451} & \cellcolor{E3CellSoft}\textbf{90.2} & \cellcolor{E3CellSoft}\textbf{78.0} & \cellcolor{E3CellSoft}\textbf{48.5} \\
GPT & 2,137 & 1,296 & 1,165 & 74.7 & 60.6 & 15.1 \\
Claude & 2,043 & 1,318 & 1,237 & 73.1 & 58.8 & 18.0 \\
\bottomrule
\end{tabular}
\caption{Per-source coverage over the full union of 4,598 judged issues. \textit{Caught}, \textit{Partial}, \textit{Missed} are raw row counts. \textit{Hit \%} is partial-inclusive recall $R_{\mathrm{hit}}$; \textit{Weighted score} is $100\cdot C_w$ (partial catch counts half); \textit{Best-rigour \%} is the share of rows where the judge ranked this source's treatment as the most thorough. The leading value in each column is in \textbf{bold}; the \system{} row is shaded for visual anchoring.}
\label{tab:method-summary}
\end{table}

\subsection{Severity-Stratified Recall}
\begin{figure}[!ht]
\centering
\includegraphics[width=\linewidth]{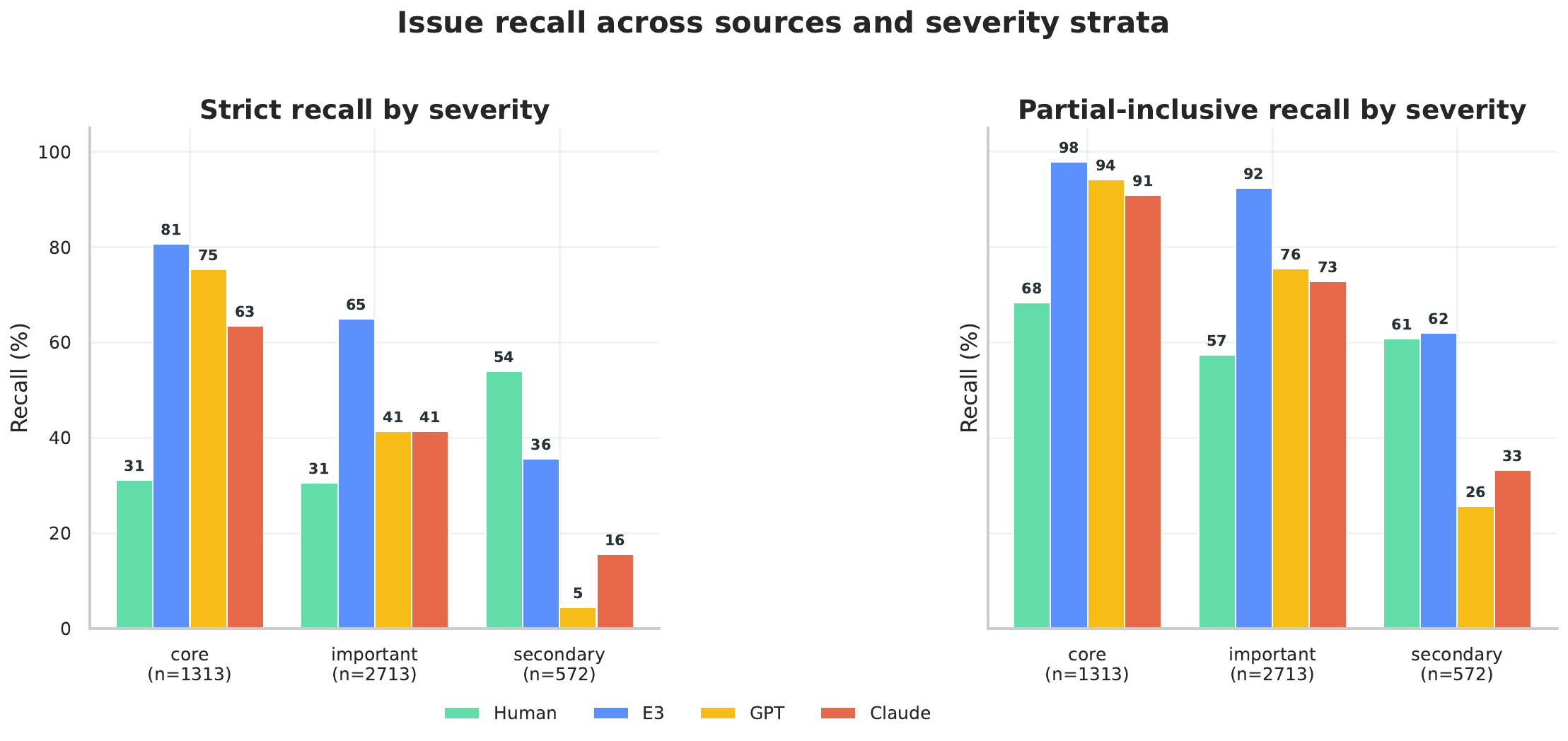}
\caption{\textbf{Strict and partial-inclusive recall by severity.} \textit{Left}: $R_{\mathrm{strict}}$, the share judged \caught{}. \textit{Right}: $R_{\mathrm{hit}}$, the share judged \caught{} or \partialhit{}. Within each severity bin bars go left-to-right Human, \system{}, GPT, Claude; values are printed above each bar, the bin size is on the $x$-axis tick. This is the paper's main side-by-side comparison: it lets the reader verify the headline numbers separately on core, important, and secondary rows. \textit{Numerical companion}: Table~\ref{tab:severity-recall}.}
\label{fig:severity-recall}
\end{figure}

An average can hide a weak spot, so we re-run the comparison inside each severity stratum. The judge labels every issue as one of three severities. \emph{Core} issues threaten the paper's main contribution; \emph{important} issues materially affect a claim; \emph{secondary} issues are stylistic, presentational, or minor. Figure~\ref{fig:severity-recall} shows strict (left) and partial-inclusive (right) recall at each stratum; Table~\ref{tab:severity-recall} prints the exact numbers.

\textbf{On the 1,313 core rows (the slice that drives accept/reject decisions), \system{} leads by \deltaup{3.7} over the next-best source}: \system{} \textbf{97.9\%}, GPT 94.2\%, Claude 90.9\%, Human 68.4\%. On the 2,713 important rows the same ordering holds (\system{} \textbf{92.4\%} \deltaup{16.8} over the next-best source; GPT 75.5\%, Claude 72.9\%, Human 57.5\%). Only the 572 secondary rows behave differently: automated recall drops (\system{} 62.1\%, GPT 25.7\%, Claude 33.2\%) while humans hold at 60.8\%. Secondary issues are typos, wording, and presentation, a category humans police well and that a ``find decision-relevant flaws'' prompt actively de-prioritises.

\begin{table}[!ht]
\centering
\small
\begin{tabular}{lrrrrr}
\toprule
\textbf{Severity} & \textbf{Issues} & \textbf{Human} & \cellcolor{E3HeadSoft}\textbf{E3} & \textbf{GPT} & \textbf{Claude} \\
\midrule
Core & 1,313 & 31.2 / 68.4 & \cellcolor{E3CellSoft}\textbf{80.7 / 97.9} & 75.4 / 94.2 & 63.4 / 90.9 \\
Important & 2,713 & 30.5 / 57.5 & \cellcolor{E3CellSoft}\textbf{64.9 / 92.4} & 41.3 / 75.5 & 41.3 / 72.9 \\
Secondary & 572 & 54.0 / 60.8 & \cellcolor{E3CellSoft}\textbf{35.7 / 62.1} & 4.5 / 25.7 & 15.6 / 33.2 \\
\bottomrule
\end{tabular}
\caption{Per-source recall stratified by judge-assigned severity, reported as \textit{strict \% / partial-inclusive \%}. Down a column: how one source's recall changes as issues become more central. Across a row: how the four sources compare at one severity. The leading source per row is in \textbf{bold}; the \system{} column is shaded.}
\label{tab:severity-recall}
\end{table}

\subsection{Recall by Severity and Decision Stratum, All Four Sources}
\begin{figure}[!ht]
\centering
\includegraphics[width=\linewidth]{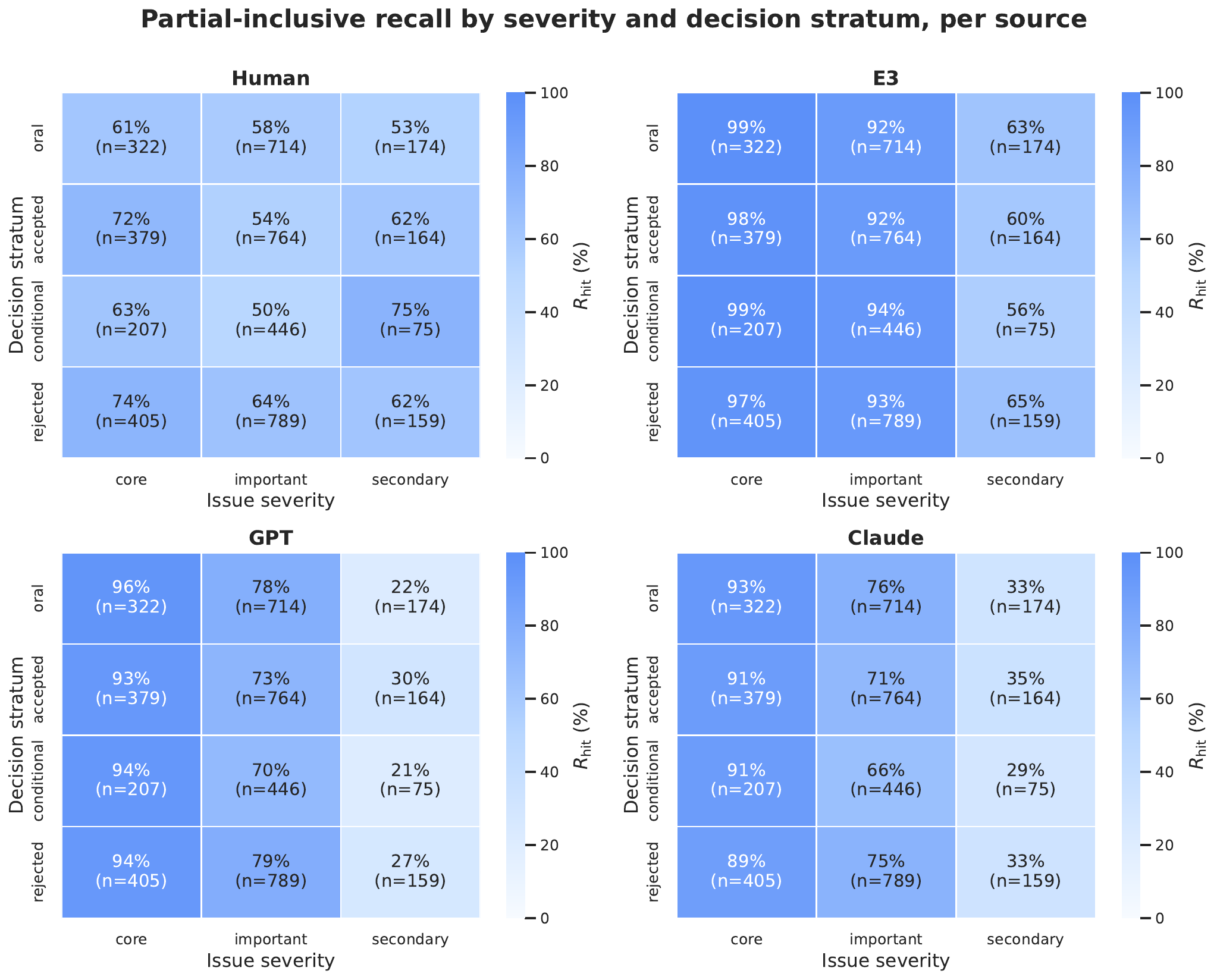}
\caption{\textbf{Partial-inclusive recall as a function of severity (columns) and decision stratum (rows), per source.} Each of the four panels uses the same axes and colour scale (0--100\%); the legend printed against the rightmost panel applies to all four. Cells annotate $R_{\mathrm{hit}}$ and the cell's sample size. Reading the panels in parallel shows that the source ordering observed in Figure~\ref{fig:coverage} is preserved at the cell level: \system{} stays above 90\% on every core cell and above 85\% on every important cell, GPT and Claude track behind at similar levels to each other, and the Human source is materially lower outside the secondary column. \textit{Numerical companion}: Table~\ref{tab:decision-summary}.}
\label{fig:severity-decision-grid}
\end{figure}

Figure~\ref{fig:severity-decision-grid} varies severity (columns) and decision stratum (rows) inside one small-multiples panel per source. The rejected stratum is the toughest test because its issue unions are larger and more varied; the core column is the most consequential because those issues drive the decision. Each cell prints its sample size, so small slices can be discounted on sight.

\begin{table}[!ht]
\centering
\small
\begin{tabular}{lrrrrrr}
\toprule
\textbf{Decision} & \textbf{Papers} & \textbf{Issues} & \textbf{Human} & \cellcolor{E3HeadSoft}\textbf{E3} & \textbf{GPT} & \textbf{Claude} \\
\midrule
Oral & 28 & 1,210 & 33.5 / 58.1 & \cellcolor{E3CellSoft}\textbf{66.7 / 89.5} & 47.3 / 74.6 & 46.0 / 74.6 \\
Accepted & 28 & 1,307 & 33.1 / 60.3 & \cellcolor{E3CellSoft}\textbf{65.1 / 89.4} & 46.3 / 73.8 & 44.4 / 72.5 \\
Conditional & 15 & 728 & 30.8 / 56.0 & \cellcolor{E3CellSoft}\textbf{63.6 / 91.5} & 44.9 / 71.8 & 42.0 / 69.4 \\
Rejected & 29 & 1,353 & 35.8 / 67.0 & \cellcolor{E3CellSoft}\textbf{66.7 / 90.8} & 46.8 / 77.0 & 44.3 / 74.3 \\
\bottomrule
\end{tabular}
\caption{Per-source recall stratified by ICLR 2026 decision folder, reported as \textit{strict \% / partial-inclusive \%}. \textit{Papers} and \textit{Issues} give the stratum size. Down a column tracks how one source's recall changes as the stratum becomes harder (oral to rejected); across a row compares the four sources at fixed decision difficulty. The leading source per row is in \textbf{bold}; the \system{} column is shaded.}
\label{tab:decision-summary}
\end{table}

\section{Human-Review Recall and Alignment}\label{sec:alignment}
Downstream users ultimately judge a critique by how well it tracks what a human reviewer would have said. Section~\ref{sec:metrics} defined the relevant quantities. The \emph{human-salient} slice is the 2,805 of 4,598 union rows that the human reviewers caught or partially caught. The \emph{human-missed} slice is the 1,793 rows they did not. Two numbers follow:
\begin{itemize}\itemsep0.2em
\item $R_{M|H}$: the share of human-salient issues source $M$ also catches; the operational definition of \emph{agreement with humans}.
\item $D_{M,\bar H}$: the count of human-missed issues source $M$ catches anyway; the operational definition of \emph{value beyond humans}.
\end{itemize}

On the human-salient slice (Table~\ref{tab:human-alignment}), \textbf{\system{} reaches 89.6\% partial-inclusive and 63.7\% strict recall}, a \deltaup{-10.4} margin over the next-best source. GPT reaches 78.6\% / 49.1\% and Claude reaches 76.8\% / 44.9\%, so the ranking from Figure~\ref{fig:coverage} holds on the human-salient subset. On the human-missed slice the Human source is $0$\% by definition and the automated sources still hit a clear majority of admitted concerns.

\begin{table}[!ht]
\centering
\small
\begin{tabular}{lrrrrr}
\toprule
\textbf{Issue stratum} & \textbf{Issues} & \textbf{Human} & \cellcolor{E3HeadSoft}\textbf{E3} & \textbf{GPT} & \textbf{Claude} \\
\midrule
Human salient (caught + partial) & 2,805 & \textbf{55.1 / 100.0} & \cellcolor{E3CellSoft}63.7 / 89.6 & 49.1 / 78.6 & 44.9 / 76.8 \\
Human missed & 1,793 & 0.0 / 0.0 & \cellcolor{E3CellSoft}\textbf{69.0 / 91.2} & 42.4 / 68.5 & 43.7 / 67.4 \\
All judged issues & 4,598 & 33.6 / 61.0 & \cellcolor{E3CellSoft}\textbf{65.8 / 90.2} & 46.5 / 74.7 & 44.4 / 73.1 \\
\bottomrule
\end{tabular}
\caption{Per-source recall on three slices of the union, reported as \textit{strict \% / partial-inclusive \%}. \textit{Strict} counts only \caught{} rows; \textit{partial-inclusive} counts \caught{} or \partialhit{}. The strata are \textit{human salient} (issues humans caught or partially caught), \textit{human missed} (the remainder of the union), and \textit{all judged issues}. On the human-salient row the Human partial-inclusive value is 100\% by construction; the strict value is the share of those rows the judge graded \caught{} rather than \partialhit{}. The leading source per row is in \textbf{bold}; the \system{} column is shaded.}
\label{tab:human-alignment}
\end{table}

\begin{figure}[!ht]
\centering
\includegraphics[width=\linewidth]{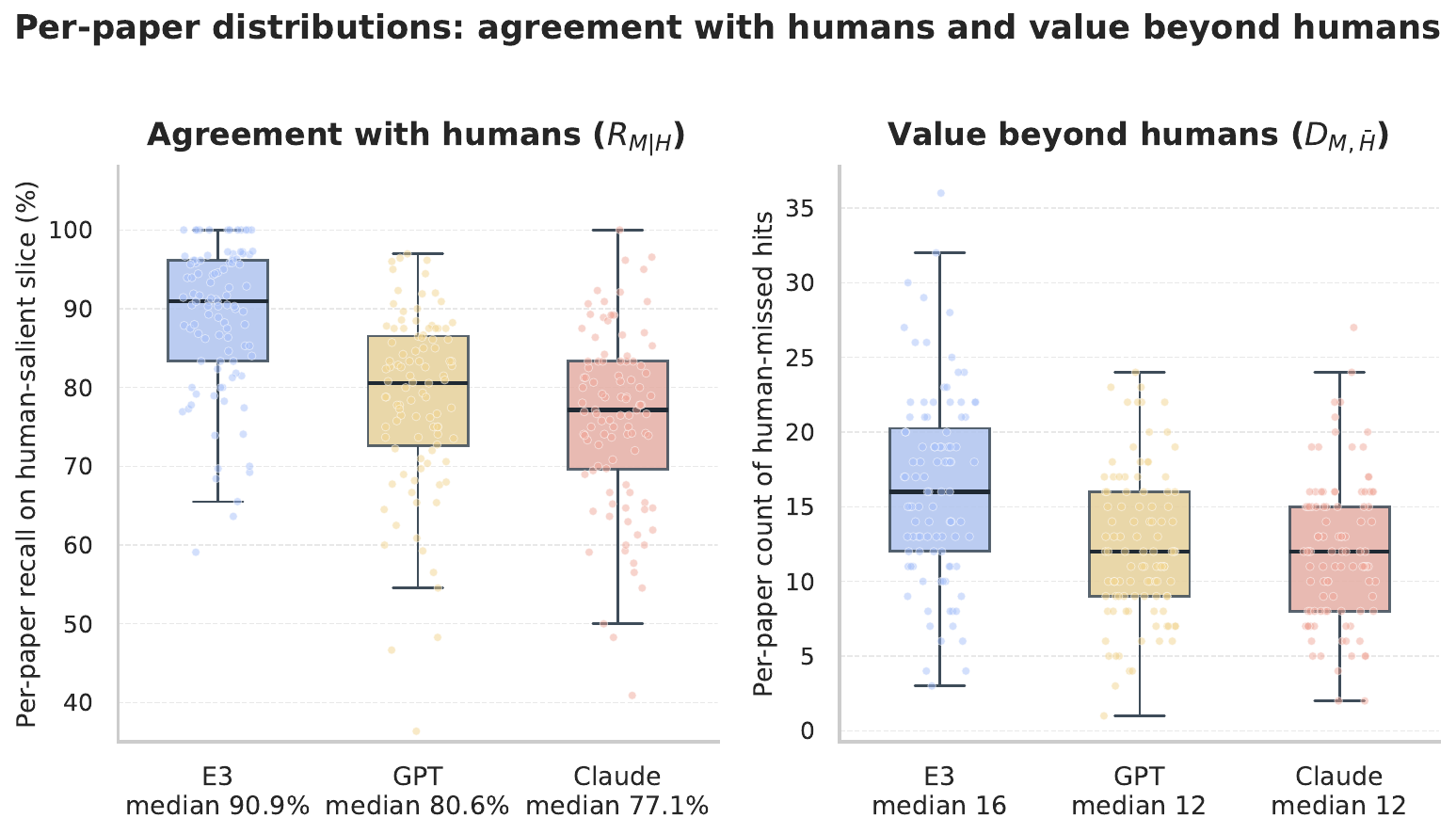}
\caption{\textbf{Per-paper distributions of agreement with humans ($R_{M|H}$, left) and value beyond humans ($D_{M,\bar H}$, right).} Each box summarises 100 papers for one source; the dots are individual papers (jittered horizontally). \textit{Left}: per-paper recall on the human-salient slice; \system{}'s box is shifted right of GPT's and Claude's, and its tails are tighter. \textit{Right}: per-paper count of human-missed issues that the source still hits; \system{} sits above GPT and Claude on both the median and the upper quartile. Median values are printed under each box. \textit{Numerical companion}: Table~\ref{tab:human-alignment}.}
\label{fig:alignment}
\end{figure}

Aggregate numbers can mask per-paper variation: a source might be aligned on some papers and complementary on others. Figure~\ref{fig:alignment} shows the per-paper distribution of both quantities, one box per source. The left panel shows the distribution of $R_{M|H}$ across the 100 papers (one point per paper, jittered); the right panel shows the distribution of $D_{M,\bar H}$. We omit the Human source because by construction $R_{H|H}=100$ and $D_{H,\bar H}=0$.

Counted in raw rows rather than rates, the value beyond humans is \textbf{1,635 issues for \system{}} versus 1,229 for GPT and 1,208 for Claude (a 406-row lead). Each row clears two filters (the source raised it, and the blinded judge accepted it into the union), so the count is conservative. We recommend it as the practical value-add metric because it measures concerns that would otherwise be missing from review notes.

\section{Taxonomy and Error Analysis}\label{sec:taxonomy}
\begin{figure}[!ht]
\centering
\includegraphics[width=0.9\linewidth]{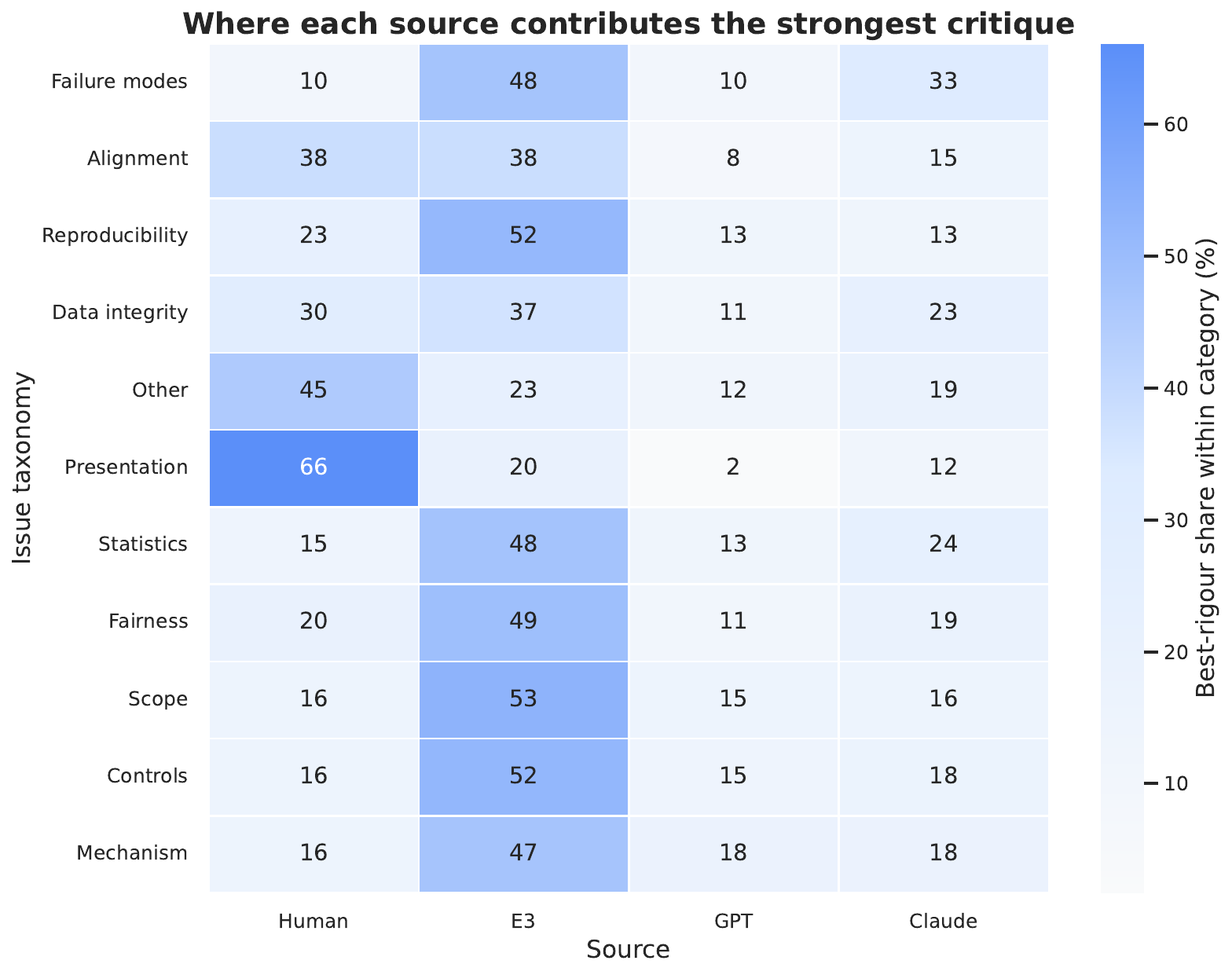}
\caption{\textbf{Best-rigour share inside each issue category.} Rows are the taxonomy buckets, columns are the four sources, and each cell is the percentage of issues in that bucket where the judge said this source gave the most evidence-backed treatment. Rows sum to 100\%. GPT and Claude take non-trivial cells in categories where they do not lead on overall recall, so they contribute additional coverage. \textit{Numerical companion}: Table~\ref{tab:taxonomy-summary}.}
\label{fig:taxonomy-heatmap}
\end{figure}

We tag every issue with one of ten named categories or a residual \emph{Other} bucket; assignment uses fixed keyword rules, so the labels are auditable rather than learned. The four largest buckets are Mechanism (1662, 36.1\%), Controls (1013, 22.0\%), Scope (822, 17.9\%), Fairness (625, 13.6\%). A high average can still hide a weak category, so Figure~\ref{fig:taxonomy-heatmap} computes best-rigour share within each bucket and lets the reader see which kind of issue each source treats most thoroughly.

\begin{table}[!ht]
\centering
\small
\begin{tabular}{lrrrrrr}
\toprule
\textbf{Taxonomy} & \textbf{Issues} & \textbf{Core \%} & \textbf{Human hit} & \cellcolor{E3HeadSoft}\textbf{E3 hit} & \textbf{GPT hit} & \textbf{Claude hit} \\
\midrule
Mechanism & 1,662 & 42.8 & 63.6 & \cellcolor{E3CellSoft}\textbf{93.0} & 83.8 & 79.1 \\
Controls & 1,013 & 21.6 & 64.9 & \cellcolor{E3CellSoft}\textbf{93.1} & 76.6 & 76.0 \\
Scope & 822 & 20.9 & 48.3 & \cellcolor{E3CellSoft}\textbf{92.3} & 72.3 & 68.7 \\
Fairness & 625 & 19.7 & 63.0 & \cellcolor{E3CellSoft}\textbf{86.6} & 65.8 & 72.5 \\
Statistics & 121 & 19.8 & 56.2 & \cellcolor{E3CellSoft}\textbf{93.4} & 76.9 & 72.7 \\
Presentation & 118 & 12.7 & \textbf{81.4} & \cellcolor{E3CellSoft}56.8 & 37.3 & 35.6 \\
Other & 88 & 13.6 & 56.8 & \cellcolor{E3CellSoft}\textbf{60.2} & 39.8 & 47.7 \\
Data integrity & 71 & 25.4 & 57.7 & \cellcolor{E3CellSoft}\textbf{81.7} & 53.5 & 56.3 \\
Reproducibility & 31 & 3.2 & 54.8 & \cellcolor{E3CellSoft}\textbf{87.1} & 51.6 & 48.4 \\
Alignment & 26 & 26.9 & 69.2 & \cellcolor{E3CellSoft}\textbf{73.1} & 57.7 & 46.2 \\
Failure modes & 21 & 52.4 & 47.6 & \cellcolor{E3CellSoft}\textbf{100.0} & 85.7 & 90.5 \\
\bottomrule
\end{tabular}
\caption{Issue-taxonomy buckets sorted by frequency. \textit{Core \%} is the share of each bucket judged core severity. Each per-source \textit{hit} column reports partial-inclusive recall ($R_{\mathrm{hit}}$) on the bucket. Reading across a row compares the four sources on issues of one topic; rows where the \textit{Human hit} cell is low while automated cells are high mark categories where automated review adds the most. The leading source per row is in \textbf{bold}; the \system{} column is shaded.}
\label{tab:taxonomy-summary}
\end{table}

\begin{figure}[!ht]
\centering
\includegraphics[width=\linewidth]{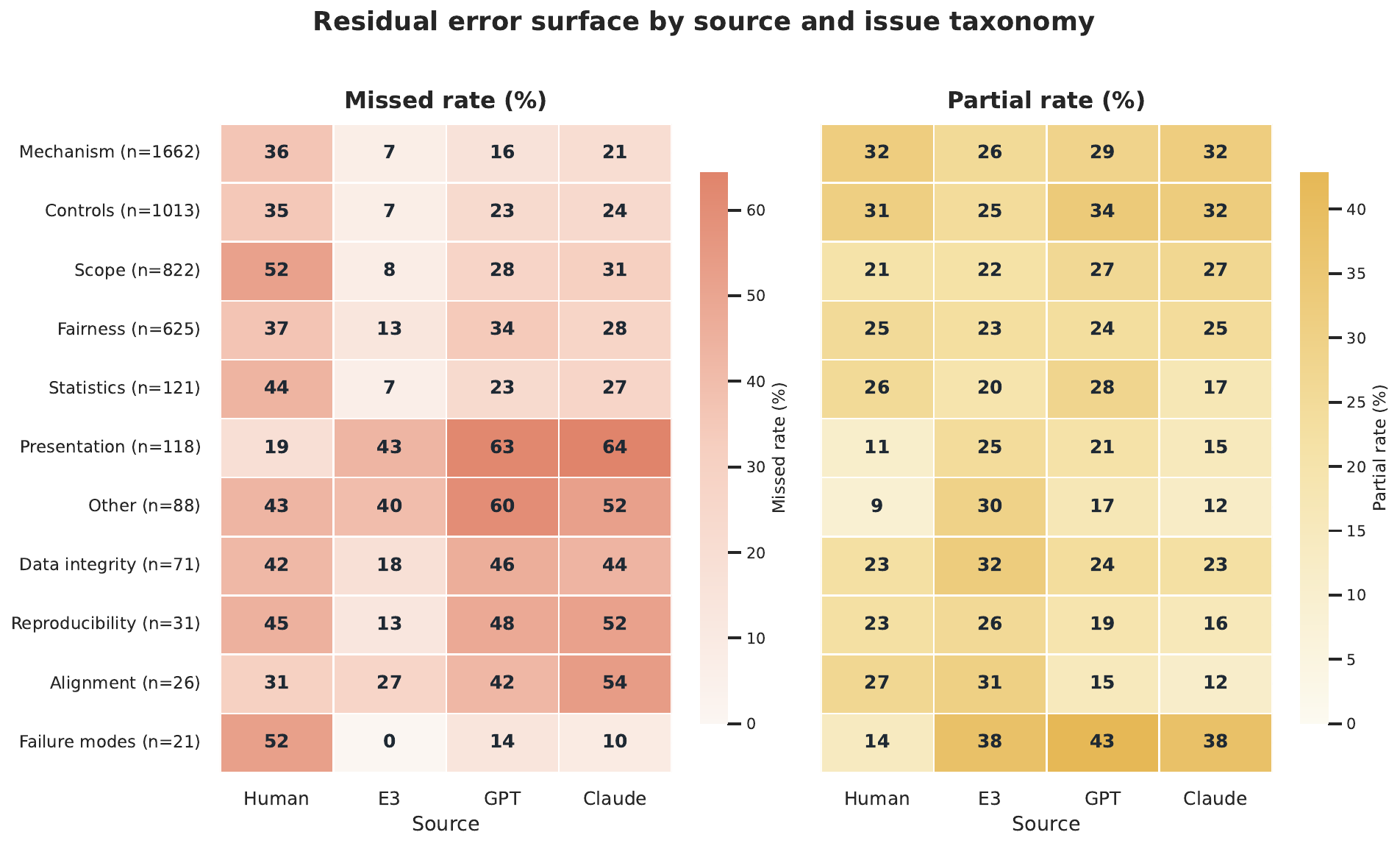}
\caption{\textbf{Residual error surface by taxonomy and source.} Rows are issue-taxonomy buckets (size in parentheses); columns are the four sources; the two panels share row labels. \textit{Left}: percentage of issues the source \missed{} entirely (soft coral). \textit{Right}: percentage the source caught only \partialhit{}ly (soft amber). Lighter cells indicate better behaviour in each panel. \system{} is the lightest column in the \missed{} panel across all rows; its non-zero residual concentrates in the \partialhit{} panel, meaning concerns are raised but not fully developed. \textit{Numerical companion}: Table~\ref{tab:taxonomy-summary}.}
\label{fig:error-surface}
\end{figure}

Figure~\ref{fig:error-surface} splits the residual-error surface into two side-by-side heatmaps. The left panel shows the \missed{} rate (the source did not address the issue at all); the right panel shows the \partialhit{} rate (the source addressed it incompletely). Together they distinguish ``absent'' from ``shallow''. The picture aligns with the recall analysis: GPT and Claude carry the darkest \missed{} cells in \textit{Other}, \textit{Failure modes / sensitivity}, and \textit{Practical impact}; \system{} stays in the lightest band across every row of the \missed{} panel. The \partialhit{} panel shows where \system{}'s errors live: it tends to mention these residual issues partially rather than miss them outright.

\section{Complementarity}\label{sec:complementarity}
\begin{figure}[!ht]
\centering
\includegraphics[width=\linewidth]{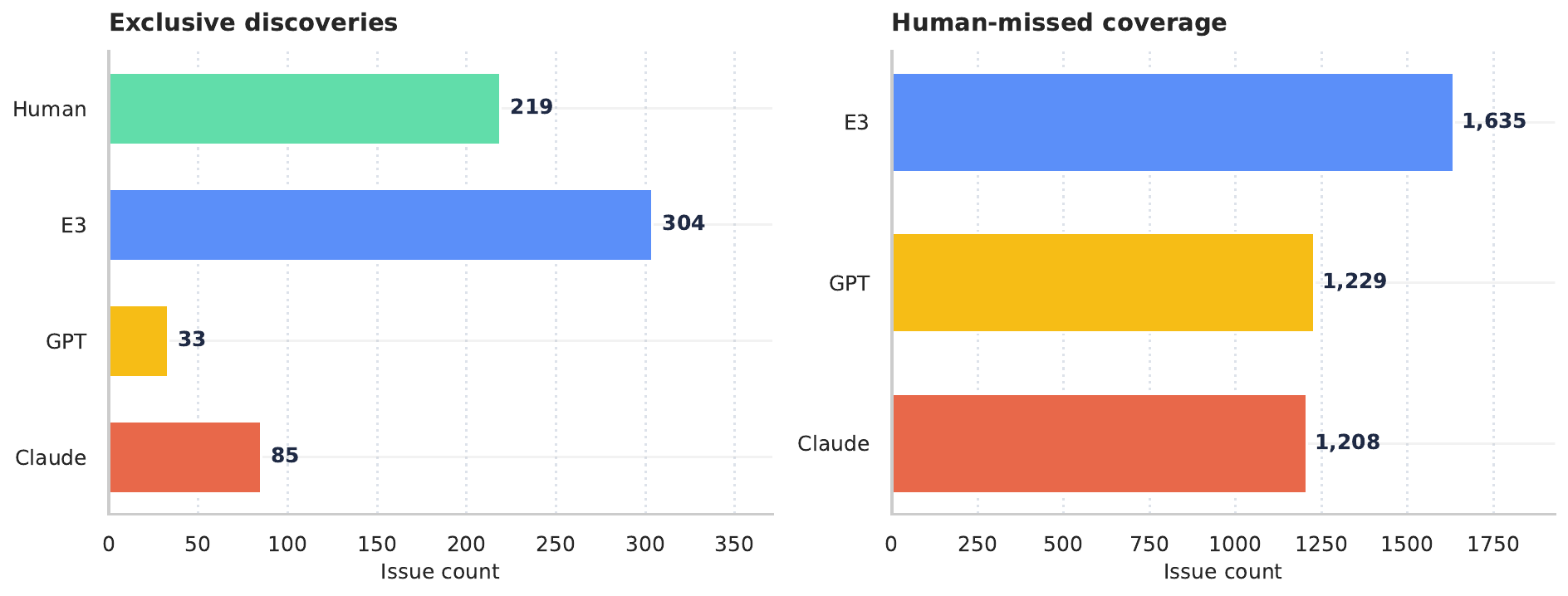}
\caption{\textbf{Complementarity by source.} \textit{Left: Exclusive discoveries}: issues caught by exactly one source and missed by all three others (raw counts). \textit{Right: Human-missed coverage}: issues each automated source caught while the human reviewers missed them; Human is omitted because a reviewer cannot catch what they missed. Bars are coloured by source for direct comparison with earlier figures. Raw counts are used because the operational quantity is the absolute volume of additional concerns each source brings to a review.}
\label{fig:unique}
\end{figure}

Two questions remain. Which source caught issues no other source caught? Which source covered the largest share of the blind spots left by the human reviewers? Figure~\ref{fig:unique} answers both in a two-panel layout: the left panel counts \emph{exclusive discoveries} (issues only one source caught), the right panel counts \emph{human-missed coverage} for the three automated sources (Human is omitted by construction). \system{} contributes the largest counts in both panels.
\begin{itemize}\itemsep0.2em
\item A \emph{unique hit} is an issue where exactly one source was \caught{} or \partialhit{}; remove that source from the protocol and the row vanishes from the recovered union.
\item A \emph{human-missed hit} is an issue the source caught while the human reviewers missed it, i.e., the source's contribution beyond what the human reviews already covered.
\end{itemize}

\clearpage
\section{Discussion}
Two conclusions follow from the measurements above.

First, \textbf{\system{} leads on every aggregate metric reported in this study}: partial-inclusive recall (\deltaup{15.5} over the next-best source), strict recall, weighted coverage, and best-rigour share (\deltaup{29.9}). The leading position holds inside every severity bin and every decision stratum (Table~\ref{tab:severity-recall}, Table~\ref{tab:decision-summary}) and inside both the human-salient and the human-missed slices (Table~\ref{tab:human-alignment}). Differences to the next-best non-\system{} source range from \deltaup{15.5} to \deltaup{29.2} on $R_{{\mathrm{{hit}}}}$ within this corpus.

Second, the GPT and Claude baselines contribute non-trivial coverage. Figure~\ref{fig:taxonomy-heatmap} shows both receiving the best-rigour tag in categories where they did not lead on overall recall, and Figure~\ref{fig:unique} shows both adding unique-hit and human-missed-hit counts. Ensembling at the issue level could therefore recover a wider union.

\section{Limitations}\label{sec:limitations}
Four limitations are worth naming explicitly. \textit{(i)} The union depends on the judge model; best-rigour attribution in particular is more subjective than \caught{}/\missed{} assignment. \textit{(ii)} The meta-judge is itself a language model: M1--M4 blinding removes name-based bias, but residual stylistic or terminological affinity between the judge and any individual LLM-based source cannot be ruled out \citep{zheng2023mtbench}. \textit{(iii)} The taxonomy is keyword-based, so it is auditable but coarse. \textit{(iv)} The corpus is ICLR 2026 papers; results may not transfer to other venues or fields.

\section{Conclusion}
Across 4,598 judged issue rows from 100 ICLR 2026 papers, \textbf{\system{} is the highest-recall source on every aggregate metric we report}: \textbf{90.2\%} partial-inclusive recall (\deltaup{15.5} over the next-best source), \textbf{65.8\%} strict recall, \textbf{48.5\%} best-rigour share (\deltaup{29.9}), \textbf{89.6\%} agreement with humans, and \textbf{1,635} human-missed concerns recovered (406 above the next-best source). The wider contribution is the protocol itself: an issue-level backtest that is transparent, reproducible, and applicable to any review system whose only observable is the critique text.

\bibliography{references}

@article{cortes2021inconsistency,
  title   = {Inconsistency in Conference Peer Review: Revisiting the 2014 {NeurIPS} Experiment},
  author  = {Cortes, Corinna and Lawrence, Neil D.},
  journal = {arXiv preprint arXiv:2109.09774},
  year    = {2021},
  url     = {https://arxiv.org/abs/2109.09774}
}

@article{beygelzimer2023consistency,
  title   = {Has the Machine Learning Review Process Become More Arbitrary as the Field Has Grown? The {NeurIPS 2021} Consistency Experiment},
  author  = {Beygelzimer, Alina and Dauphin, Yann N. and Liang, Percy and Vaughan, Jennifer Wortman},
  journal = {arXiv preprint arXiv:2306.03262},
  year    = {2023},
  url     = {https://arxiv.org/abs/2306.03262}
}

@article{kapoor2023leakage,
  title   = {Leakage and the Reproducibility Crisis in Machine-Learning-Based Science},
  author  = {Kapoor, Sayash and Narayanan, Arvind},
  journal = {Patterns},
  volume  = {4},
  number  = {9},
  pages   = {100804},
  year    = {2023},
  doi     = {10.1016/j.patter.2023.100804}
}

@inproceedings{liang2024monitoring,
  title     = {Monitoring {AI}-Modified Content at Scale: A Case Study on the Impact of {ChatGPT} on {AI} Conference Peer Reviews},
  author    = {Liang, Weixin and Izzo, Zachary and Zhang, Yaohui and Lepp, Haley and Cao, Hancheng and Zhao, Xuandong and Chen, Lingjiao and Ye, Haotian and Liu, Sheng and Huang, Zhi and McFarland, Daniel A. and Zou, James},
  booktitle = {Proceedings of the 41st International Conference on Machine Learning (ICML)},
  year      = {2024},
  url       = {https://proceedings.mlr.press/v235/liang24b.html}
}

@inproceedings{zheng2023mtbench,
  title     = {Judging {LLM-as-a-Judge} with {MT-Bench} and {Chatbot Arena}},
  author    = {Zheng, Lianmin and Chiang, Wei-Lin and Sheng, Ying and Zhuang, Siyuan and Wu, Zhanghao and Zhuang, Yonghao and Lin, Zi and Li, Zhuohan and Li, Dacheng and Xing, Eric P. and Zhang, Hao and Gonzalez, Joseph E. and Stoica, Ion},
  booktitle = {Advances in Neural Information Processing Systems (NeurIPS), Datasets and Benchmarks Track},
  year      = {2023},
  url       = {https://arxiv.org/abs/2306.05685}
}

@article{white2024livebench,
  title   = {{LiveBench}: A Challenging, Contamination-Free {LLM} Benchmark},
  author  = {White, Colin and Dooley, Samuel and Roberts, Manley and Pal, Arka and Feuer, Ben and Jain, Siddhartha and Shwartz-Ziv, Ravid and Jain, Neel and Saifullah, Khalid and Naidu, Siddartha and Hegde, Chinmay and LeCun, Yann and Goldstein, Tom and Neiswanger, Willie and Goldblum, Micah},
  journal = {arXiv preprint arXiv:2406.19314},
  year    = {2024},
  url     = {https://arxiv.org/abs/2406.19314}
}

\appendix
\section{Exact LLM Baseline Prompt}
\label{app:vanilla-prompt}
The following text is generated directly from the baseline source code. Runtime placeholders are filled with the paper title and extracted paper text.
\begin{footnotesize}
\begin{verbatim}
SYSTEM PROMPT
=========
You are a rigorous ML/AI research reviewer with deep subfield expertise. You are fair, 
  technical, and evidence-driven. Your task is simple: find the most important flaws, risks,
   unsupported claims, missing validations, and domain-level issues in the paper. Do not be 
  hostile for the sake of it. Do not assume generic benchmark criticisms are sufficient; 
  actively consider domain-specific concerns that a specialist would raise. Support your 
  points with concrete evidence from the paper when possible.

USER PROMPT TEMPLATE
==================
Review this ML/AI paper with one goal: find the most decision-relevant flaws.

Important instructions:
- Be fair, technical, and domain-aware.
- Look for domain-specific issues that a generic reviewer might miss.
- Focus on unsupported claims, hidden assumptions, weak baselines, edge cases, confounds, 
  missing validations, and incorrect or overstated statements.
- Do not use any historical review notes or external checklists.

Return markdown with these sections exactly:
- ## CORE CLAIM
- ## MAIN RISKS
- ## DOMAIN-SPECIFIC CONCERNS
- ## MISSING VALIDATION
- ## SHARPEST FLAW
- ## ACCEPTANCE RECOMMENDATION
- ## POINTERS

For ## POINTERS:
- Output 8-20 bullets.
- Each bullet must contain exactly one concrete criticism or risk.
- Do not include praise.
- Do not merge multiple issues into one bullet.

In ## ACCEPTANCE RECOMMENDATION use this exact format:
**Score:** X/10
**Decision:** Accept / Borderline / Reject / Strong Reject
**Tier prediction:** oral / accepted / conditional / rejected
**Reasoning:** <1 sentence>

Paper title: {title}

Paper text:
{paper_text}
\end{verbatim}
\end{footnotesize}

\section{Exact Blinded Judge Prompt Template}
\label{app:judge-prompt}
The following text is generated directly from the issue-matrix pipeline source code. Runtime payload fields are shown as placeholders to avoid reproducing all 100 papers and reviews inside the report.
\begin{scriptsize}
\begin{verbatim}
SYSTEM PROMPT
=========
You are a meticulous research-paper review meta-judge. You exhaustively extract 
  decision-relevant review concerns from four blinded review sources (M1, M2, M3, M4), then 
  build a normalized union-of-issues matrix. Prefer granular, non-overlapping issue rows 
  over coarse summaries. Stay domain-agnostic: apply the same standards regardless of field 
  or venue. Score sources only by content quality — do not infer provider identity from 
  labels or writing style. Return only valid JSON.

USER PROMPT TEMPLATE
==================
Create an EXHAUSTIVE judged union-of-issues matrix from these sources.

You are comparing whether each blinded source (M1, M2, M3, M4) identified each issue.
Your goal is maximum recall of distinct, decision-relevant concerns — not a short executive 
  summary.
Use ONLY the labels M1, M2, M3, M4 in your JSON — never substitute provider or method names.

━━━━━━━━━━━━━━━━━━━━━━━━━━━━━━━━━━━━━━━━
PHASE 1 — EXTRACT (do this mentally before writing JSON)
━━━━━━━━━━━━━━━━━━━━━━━━━━━━━━━━━━━━━━━━
For EACH source separately (M1, M2, M3, M4):
  1. List every distinct concern, weakness, question, or requested evidence.
  2. Include both major flaws and secondary concerns (presentation, citations, scope gaps).
  3. Do NOT merge related items yet — keep them separate at this stage.
  4. For M1: note reviewer identifiers when available and whether a meta-review or decision 
    record
     indicates the concern was later addressed.
  5. For M2/M3/M4: cite specific evidence from the review when available.


Coverage checklist — scan ALL sources for issues in each category before deduplicating:
  A. Core claims and mechanism validity (theory vs practice, internal consistency, causal 
    attribution)
  B. Method design (assumptions, design choices, implementation fidelity)
  C. Comparison fairness (baseline choice, fidelity, missing alternatives, matched 
    resources)
  D. Controls and ablations (component isolation, simpler alternatives, confound removal)
  E. Evaluation scope (datasets, settings, populations, generalization, external validity)
  F. Metrics and costs (primary vs proxy metrics, efficiency, resource use, strength of 
    claims)
  G. Statistical rigor (sample sizes, variance, uncertainty, significance, replication and 
    anything else relevant to the paper)
  H. Validity of proxies and intermediate signals (calibration, measurement, diagnostic 
    evidence)
  I. Interpretability, failure modes, and edge cases
  J. Novelty, related work, and citation accuracy
  K. Reproducibility, deployment, and practical applicability
  L. Data integrity (leakage, overlap, selection bias, annotation quality)
  M. Presentation clarity (writing, figures, tables, overclaiming)
  N. Anything else relevant to the paper.


━━━━━━━━━━━━━━━━━━━━━━━━━━━━━━━━━━━━━━━━
PHASE 2 — UNION (build the matrix rows)
━━━━━━━━━━━━━━━━━━━━━━━━━━━━━━━━━━━━━━━━
  1. Take the union of all Phase-1 items across all sources.
  2. Merge ONLY when two items are truly the same concern (same root cause, same missing 
    evidence
     or experiment). Do NOT merge merely because issues are related or thematically 
       adjacent.
     When in doubt, keep separate rows rather than combine.
  3. Include every distinct concern raised by any source. Row count should reflect source 
    breadth —
     do not collapse many distinct concerns into a small summary set.
  4. Each row must be a real decision-relevant review concern, not praise.
  5. Order rows by severity: foundational validity issues first, then experimental or 
    evidentiary
     gaps, then secondary or presentation issues.

━━━━━━━━━━━━━━━━━━━━━━━━━━━━━━━━━━━━━━━━
PHASE 3 — SCORE each source per row
━━━━━━━━━━━━━━━━━━━━━━━━━━━━━━━━━━━━━━━━
Use exactly these source statuses: "Caught", "Partial", "Missed".
Use exactly these best_rigour values: "M1", "M2", "M3", "M4".

  - "Caught"     = source clearly identified the issue with useful specificity.
  - "Partial"    = source gestured at it but missed key mechanism, evidence, or implication.
  - "Missed"     = source did not identify it.

  - best_rigour  = the single source with the most detailed, evidence-backed, actionable 
    treatment.
  - Score by content quality only. Do not favor length, verbosity, or apparent model 
    identity.
  - Notes should be 1–3 sentences — specific but not essay-length.
  - In notes, refer to sources as M1/M2/M3/M4 only (never "human", "GPT", "Claude", etc.).

Return ONLY valid JSON with this schema:
{
  "paper_title": "...",
  "decision": "...",
  "summary": {
    "total_issues": 0,
    "best_rigour_counts": {"M1": 0, "M2": 0, "M3": 0, "M4": 0},
    "short_takeaway": "..."
  },
  "issues": [
    {
      "topic": "...",
      "severity": "core|important|secondary",
      "m1": {"status": "Caught|Partial|Missed", "note": "..."},
      "m2": {"status": "Caught|Partial|Missed", "note": "..."},
      "m3": {"status": "Caught|Partial|Missed", "note": "..."},
      "m4": {"status": "Caught|Partial|Missed", "note": "..."},
      "best_rigour": "M1|M2|M3|M4"
    }
  ],
  "analysis": {
    "what_m1_added": "...",
    "what_m2_added": "...",
    "what_m3_added": "...",
    "what_m4_added": "...",
    "prompt_lessons": ["...", "..."]
  }
}

Sources:
{
  "paper_title": "{paper_title}",
  "allowed_statuses": [
    "Caught",
    "Partial",
    "Missed"
  ],
  "allowed_best_rigour": [
    "M1",
    "M2",
    "M3",
    "M4"
  ],
  "M1_reviews_and_decision_record": {
    "reviews": "{M1_public_reviews_json}",
    "meta_review": "{M1_meta_review_json}",
    "decision": "{M1_decision_json}",
    "metadata": "{M1_metadata_json}"
  },
  "M2_review_full": "{M2_review_full}",
  "M3_review_full": "{M3_review_full}",
  "M4_review_full": "{M4_review_full}"
}
\end{verbatim}
\end{scriptsize}

\end{document}